\documentclass[letterpaper,10 pt,conference]{ieeeconf}
\IEEEoverridecommandlockouts
\usepackage{graphicx}
\usepackage{cite}
\usepackage{amsmath}

\usepackage{amsfonts}
\usepackage{pifont}
\usepackage{multirow}
\usepackage{subcaption}
\usepackage{algorithmicx}
\usepackage[ruled]{algorithm}
\usepackage[noend]{algpseudocode}

\newcommand{\figlab}[1]{\label{fig:#1}}
\newcommand{\figref}[1]{Fig.~\ref{fig:#1}} 
\newcommand{\tablab}[1]{\label{tab:#1}}
\newcommand{\tabref}[1]{Table~\ref{tab:#1}} 
\newcommand{\forlab}[1]{\label{for:#1}}
\newcommand{\forref}[1]{Eqn.~(\ref{for:#1})} 
\newcommand{\algolab}[1]{\label{algorithm:#1}}
\newcommand{\algoref}[1]{Algorithm~\ref{algorithm:#1}} 

\begin{document}

\title{Component Influence-Driven Fastener Reduction for Robotic Disassemblability-Aware Design Simplification}
\author{Takuya Kiyokawa$^*$, Tomoki Ishikura$^{**}$, Shingo Hamada$^{**}$, \\ Genichiro Matsuda$^{**}$, and Kensuke Harada$^*$
\thanks{$^*$Department of Systems Innovation, Graduate School of Engineering Science, The University of Osaka, 1-3 Machikaneyama, Toyonaka, Osaka, Japan.}%
\thanks{$^**$Manufacturing Innovation Division, Panasonic Holdings Corporation Kadoma, Osaka, Japan}%
}

\maketitle

\begin{abstract}
To accelerate automated remanufacturing, robotic disassembly must be considered during the product design phase. However, designers currently lack quantitative feedback to identify which structural elements hinder robotic operations. To address this, this study proposes an analytical framework that provides actionable redesign guidance focused on fastener reduction, as fasteners are numerous and ubiquitous components found in almost all manufactured products. Using a Computer-Aided Design (CAD) model and its automatically generated Contact-Connection-Constraint (CCC) graph, the framework translates robotic disassembly sequence planning outcomes into component influence scores. These scores reflect how often a component causes structural constraint violations or evaluation objective deteriorations in the robotic disassembly sequence. To visually highlight structural hindrances, the framework projects these scores onto the CAD geometry as 3D heatmaps. The system then analytically simulates the removal of highly influential fasteners. It reports the expected reductions in structural constraints, tool changes, and robot travel distances, while preventing structurally unsafe modifications by evaluating geometric stability metrics. Experiments on seven household appliances demonstrate that the framework successfully targets redundant fasteners. Removing the recommended fasteners simplified the structural dependencies by eliminating between 8 and 132 structural constraints on the graph depending on each product's structural configuration. Furthermore, it improved robotic operational efficiency by eliminating unnecessary tool change operations and shortening travel distances by 165 to 1675 millimeters wherever structurally permissible.
\end{abstract}


\section{Introduction}
Although it is important to automate remanufacturing processes by taking advantage of robotics technologies to drive innovations in sustainable manufacturing~\cite{Poschmann2020,Ranjan2025}, robotic remanufacturing of end-of-life consumer products requires quantitative disassembly feedback during the mechanical design process~\cite{Xu2025,Khendry2025}. Existing Design for Disassembly (DfD) frameworks traditionally focus on manual dismantling effort, structural complexity, and CAD-based geometric modifications for semi-destructive processing~\cite{Shiraishi2015,Fukushige2017}. Meanwhile, evaluations of operational difficulty in robotic manipulation~\cite{Brown2020,Kiyokawa2023RCIM} rarely connect to concrete redesign guidelines.

This study is based on the hypothesis that removing components with high influence scores in the robotic disassembly sequence reduces sequence constraints and improves operational efficiency. Based on this premise, this paper proposes an analytical framework that translates the position of a part within a feasible robotic disassembly sequence into a per-part influence score. The system calculates this score by counting how often a positional swap breaks structural constraints or degrades the evaluation objectives of a robotic disassembly sequence planner.

The proposed pipeline utilizes these scores to systematically reduce a highly actionable design parameter: the number of fasteners. Fasteners are targeted because they are the most accessible elements for design modification that directly dictate disassembly time. Given a CAD model and its corresponding Contact-Connection-Constraint (CCC) graph~\cite{Kiyokawa2025}, the system operates analytically without invoking generative models~\cite{Wu2021,Regenwetter2022,Ma2023,Li2024,Wang2025,Kim2026,Zhang2026}. By simulating the removal of high-influence fasteners and recounting the remaining graph constraints, it evaluates direct improvements in robotic operational efficiency, specifically the reduction of structural constraints, tool-change counts, and inter-part travel distances.

The primary contribution of this paper is this analytical framework, which presents these evaluations as 3D influence heatmaps and a ranked reduction candidate list, which is sorted in descending order based on the influence scores. This auditable output allows designers to carefully balance geometric stability with robotic operational efficiency, aligning with the principles of Design for Remanufacturing (DfRem) and Design for Robotic Assembly (DfRA)~\cite{Xu2025,Khendry2025}. Finally, experiments on seven household appliances validate the approach, demonstrating how the extracted information serves as a practical, quantitative redesign guideline for fastener removal without exhaustive geometric modifications.

\section{Related Works}
To design products for robotic disassembly, engineers must account for robot-specific constraints, such as kinematic limits, rigid sequence dependencies (e.g., the requirement to remove a specific exterior cover before accessing target internal parts), and tool-change overheads. Consequently, recent research in DfRA and consumer electronic redesign specifically targets the integration of these physical constraints into the mechanical design phase for automated operations~\cite{Ricard2024,Akkawi2025}. For example, iterative frameworks refine structures for physical feasibility~\cite{Khendry2025}. Meanwhile, generative approaches leverage deep learning and Large Language Models (LLMs) to explore conceptual designs and synthesize CAD geometries~\cite{Wu2021,Regenwetter2022,Ma2023,Li2024,Wang2025,Kim2026,Zhang2026}. Additionally, remanufacturing guidelines are derived from product information and lifecycle scenarios to support a circular economy~\cite{Xu2025}. While these advances provide abstract guidelines that lack specific geometrical modification steps, there remains a need for quantitative metrics that link specific structural choices to both geometric stability and robotic operational efficiency.

On the operational execution side, recent advances have established robust methods for robotic disassembly sequence planning, task allocation, and motion scheduling under structural constraints~\cite{Laili2022,Wang2022,Meng2023,Kiyokawa2025,Kiyokawa2026schedule}. The maturity of these planning frameworks now allows for the quantitative evaluation of robotic disassemblability. Rather than treating planning as a final step to adapt to a fixed product, these established technologies can serve as an analytical basis to evaluate the product design itself.

This paper bridges the gap between execution-level planning and structural design evaluation. By applying established sequence-and-motion planning capabilities, this study translates planning outcomes into structured redesign feedback. The framework distinguishes itself by transforming sequence-derived influence metrics into concrete, per-fastener reduction recommendations ranked by their combined influence scores. This provides human engineers with influence heatmaps and ranked recommendations for improving robotic disassemblability, linking geometric stability and robotic operational efficiency directly back to the mechanical design process.

\section{Methods}

\subsection{System Overview}
The proposed analytical framework evaluates product structures to generate redesign candidates for robotic disassembly. The process is summarized in \algoref{full_pipeline}, which integrates three distinct phases into a single execution flow.

\begin{algorithm}[tb]
    \caption{Redesign Recommendation Pipeline}
    \algolab{full_pipeline}
    \begin{algorithmic}[1]
        \Require CCC graph (edges $E_{CC}$, labels $\boldsymbol{I}$), matrices (constraint transition $\boldsymbol{X}_{ctm}$, contact $\boldsymbol{X}_{ctc}$), baseline sequence $S$, removal limit $R_\mathrm{max}$
        \Ensure Ranked candidate list $\mathcal{S}$ with metrics
        \Statex \textbf{// Phase 1: Influence Estimation}
        \ForAll{component $i \in \{1,\dots,n\}$}
            \State Count penalties $v_{\mathrm{const}}, v_{\mathrm{obj}}$ via sequence swaps
            \State Compute influence scores $c_{\mathrm{const}, i}, c_{\mathrm{obj}, i}$
            \State Compute combined score $s(i)$
        \EndFor
        \Statex \textbf{// Phase 2: Host-Fastener Grouping}
        \State Identify fasteners via labels $\boldsymbol{I}$; group by host part $p$ using $E_{CC}$
        \State $\mathcal{G} \gets$ Set of host groups $(p, \mathcal{F}_p)$ where $|\mathcal{F}_p| \geq 2$
        \Statex \textbf{// Phase 3: Influence-Driven Selection}
        \State $\mathcal{S} \gets \emptyset$
        \ForAll{host group $(p, \mathcal{F}_p) \in \mathcal{G}$}
            \State Sort $\mathcal{F}_p$ in descending order by $s(f)$
            \For{$r = 1$ to $\min(R_\mathrm{max}, |\mathcal{F}_p|)$}
                \State $\mathcal{R} \gets$ top-$r$ fasteners from sorted $\mathcal{F}_p$
                \State $\Delta E \gets E_{\mathrm{after}} - E_{\mathrm{before}}$ \Comment{via Eqn. (6)}
                \If{no parts are isolated after removing $\mathcal{R}$}
                    \State Compute $\rho_J, \rho_A, \Delta T, \Delta D$ \Comment{via Eqns. (7)--(10)}
                    \State $\mathcal{S} \gets \mathcal{S} \cup \{ (\mathcal{R}, \Delta E, \Delta T, \Delta D, \rho_J, \rho_A) \}$
                \EndIf
            \EndFor
        \EndFor
        \State \Return $\mathcal{S}$ sorted in descending order by the subset influence score $\sum_{f \in \mathcal{R}} s(f)$
    \end{algorithmic}
\end{algorithm}

First, the system estimates the structural influence of each component by simulating positional swaps in the baseline robotic disassembly sequence. Next, it utilizes the CCC graph (\figref{cccgraph_simple}) to identify host parts and enumerate feasible fastener-reduction candidates. Finally, it evaluates the geometric stability and robotic operational efficiency impacts of each candidate.

\begin{figure}[tb]
    \centering
    \includegraphics[width=\linewidth]{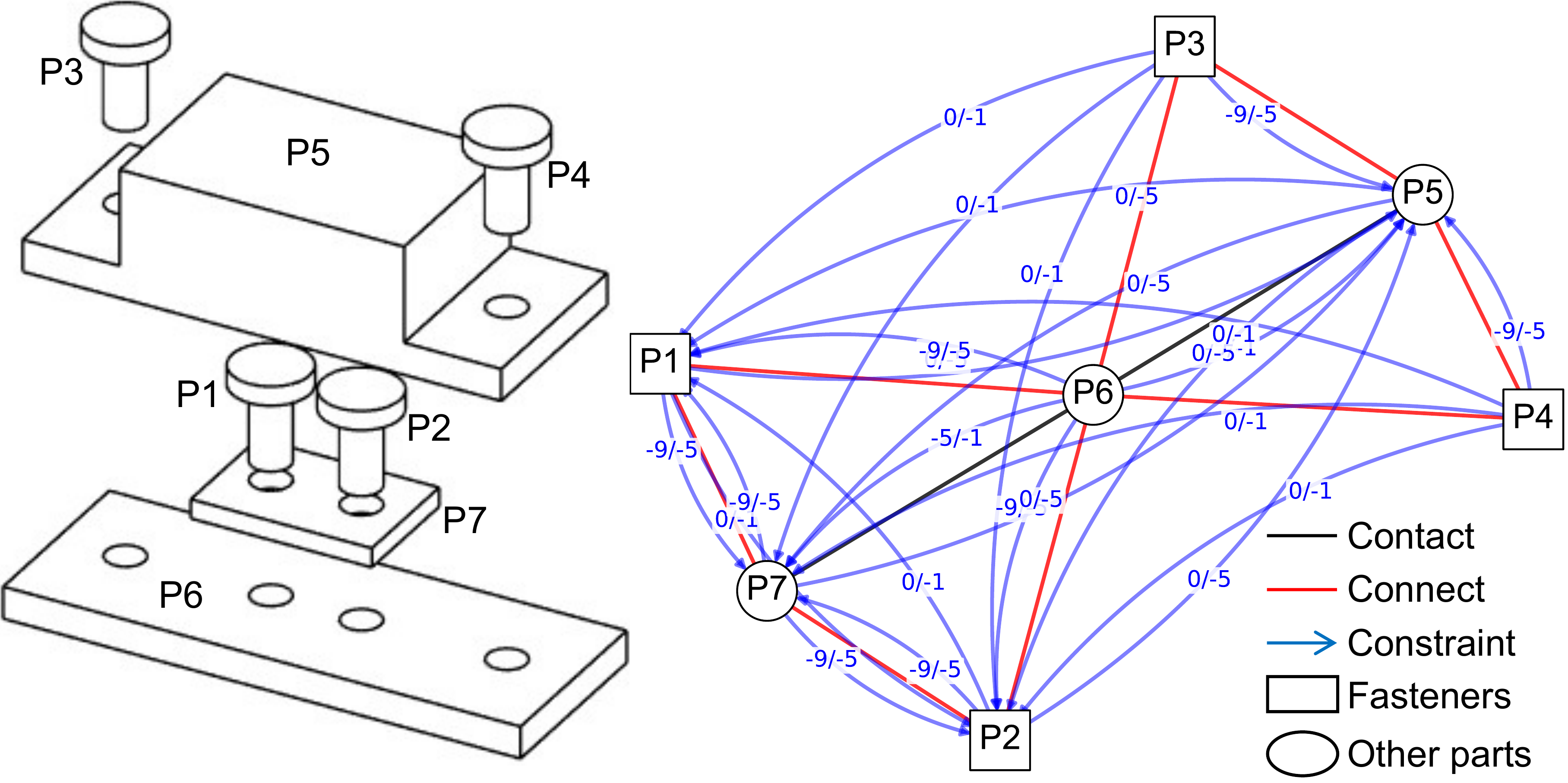}
    \caption{Illustrative CCC graph structure for a seven-part example assembly. Circles and triangles represent part and fastener nodes, respectively. Black lines indicate contact edges, red lines indicate connection edges, and blue arrows represent constraint edges. $\Delta E$ evaluates the reduction of these constraint edges.}
    \figlab{cccgraph_simple}
\end{figure}

\begin{figure*}[!tb]
    \centering
    \includegraphics[width=\linewidth]{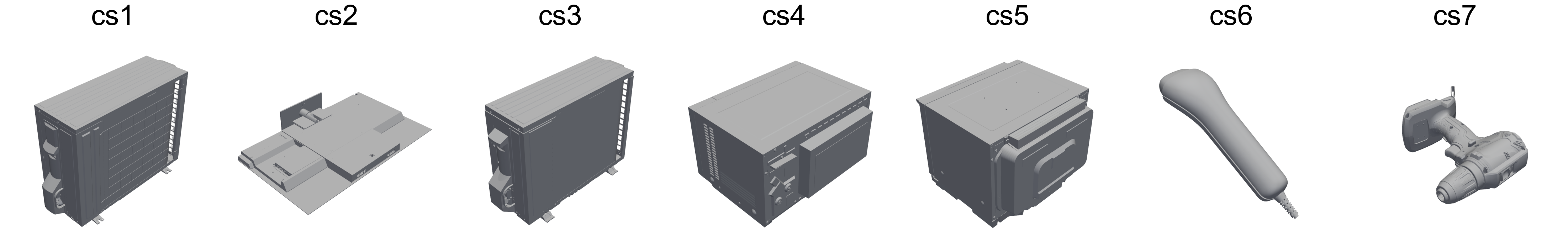}
    \caption{The seven case-study products (cs1--cs7): condenser unit, television, air-conditioner outdoor unit, microwave oven, compact microwave oven, optical beauty device, and power tool.}
    \figlab{objects}
\end{figure*}

\begin{figure}[!tb]
    \centering
    \begin{subfigure}[t]{0.49\linewidth}
        \centering
        \includegraphics[width=\linewidth]{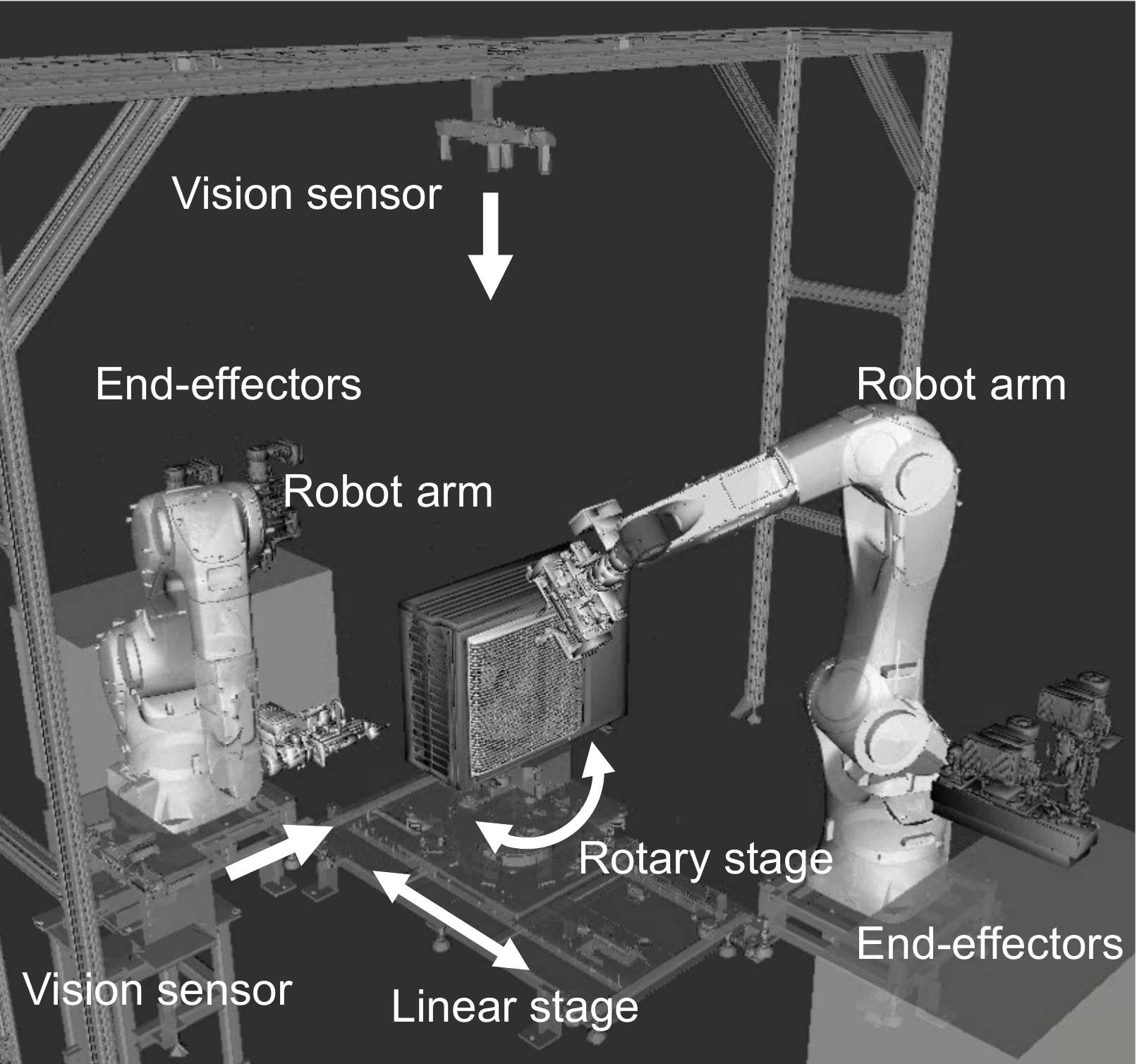}
        \caption{Large dual-arm cell.}
    \end{subfigure}\hfill
    \begin{subfigure}[t]{0.49\linewidth}
        \centering
        \includegraphics[width=\linewidth]{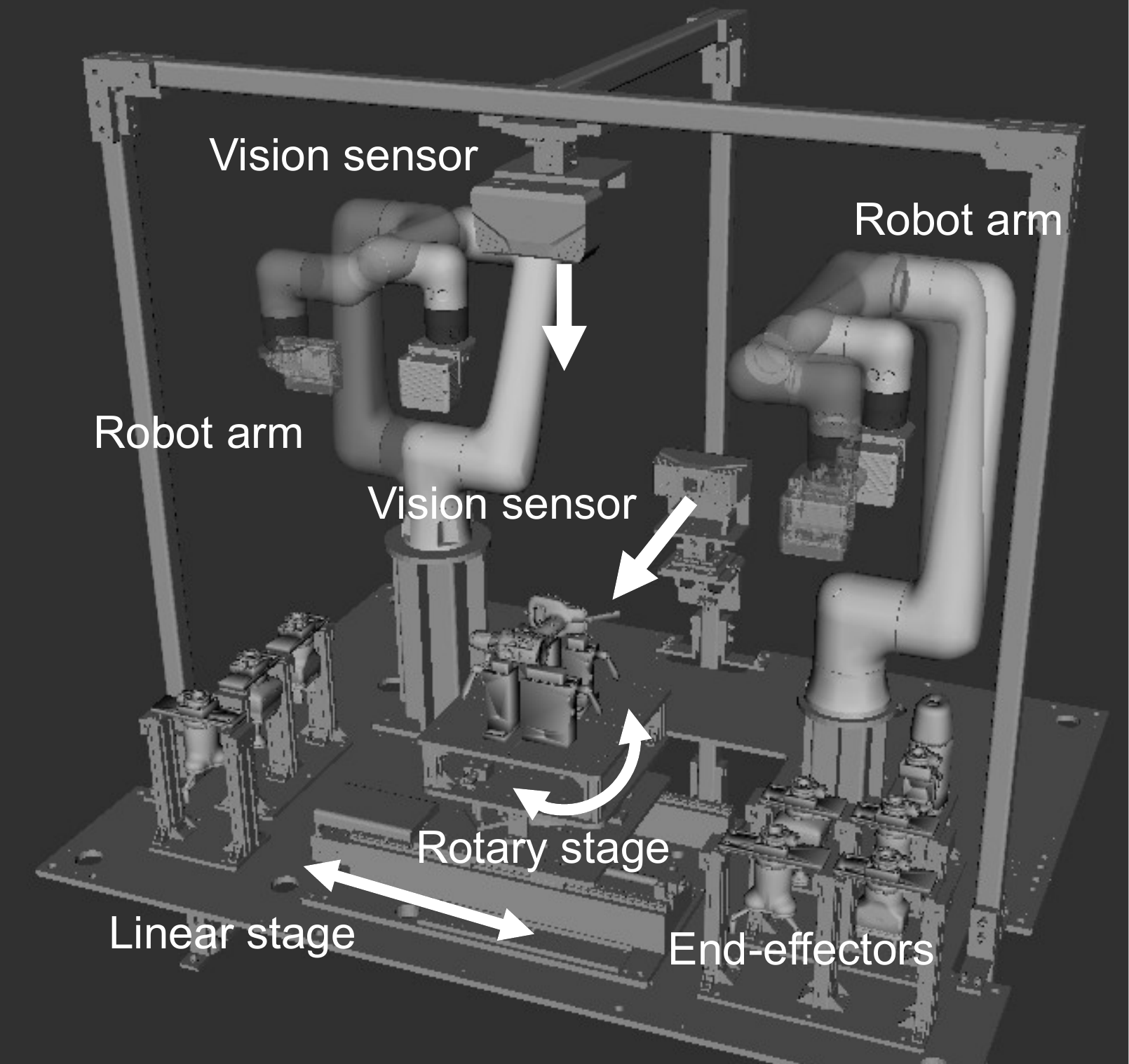}
        \caption{Small dual-arm cell.}
    \end{subfigure}
    \caption{The two robotic cells used to evaluate $\Delta T$ and $\Delta D$. Both utilize dual-arm robots equipped with multi-view vision sensors, linear and rotary stages to rotate target objects, and tool racks placed near each arm.}
    \figlab{robot_cells}
\end{figure}

\begin{figure*}[!tb]
    \centering
    \includegraphics[width=\linewidth]{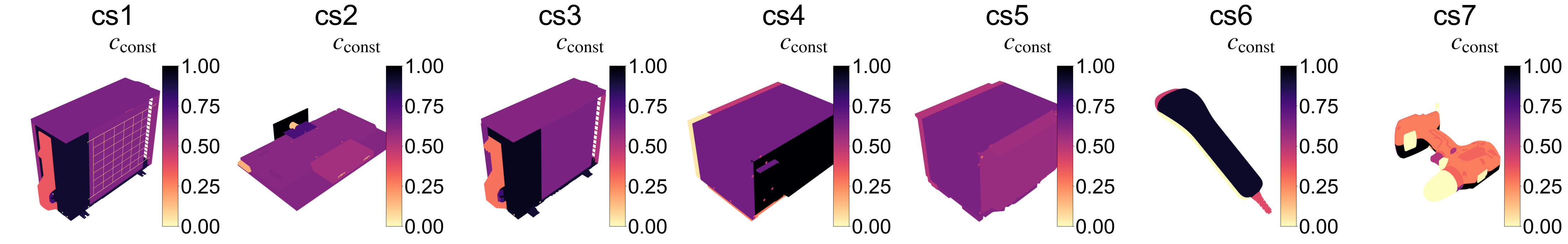}
    \caption{Assembly-level $c_\mathrm{const}$ heatmaps for the seven products. Darker colors denote high scores, indicating major structural constraints.}
    \figlab{heatmaps_const_overview}
    \includegraphics[width=\linewidth]{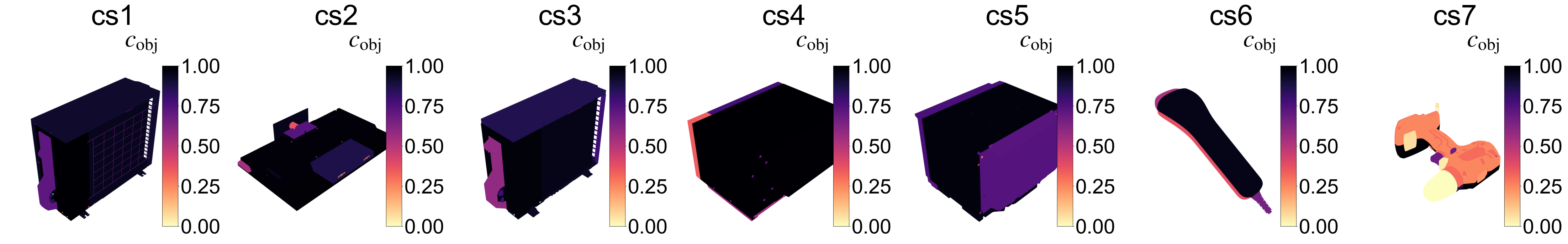}
    \caption{Assembly-level $c_\mathrm{obj}$ heatmaps for the seven products. Darker colors highlight components degrading operational efficiency.}
    \figlab{heatmaps_obj_overview}
\end{figure*}

\begin{figure*}[!tb]
    \centering
    \includegraphics[width=\linewidth]{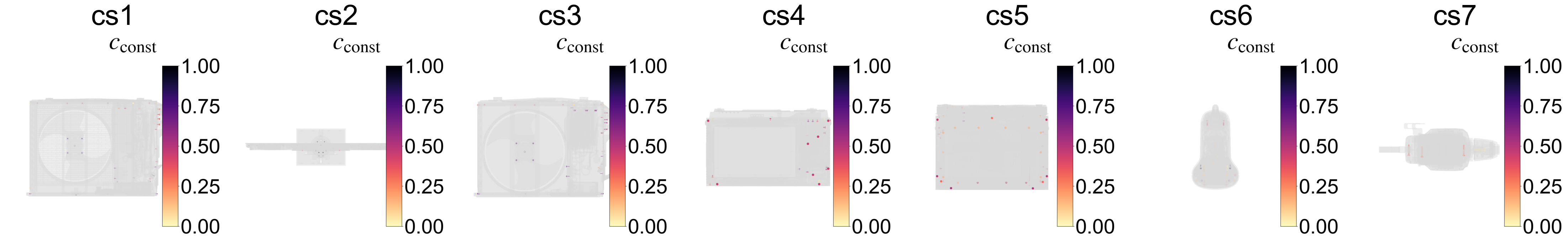}
    \caption{Top-down directive view of $c_\mathrm{const}$ identifying structurally critical fasteners clustered on upper faces.}
    \figlab{directive_topview}
    \includegraphics[width=\linewidth]{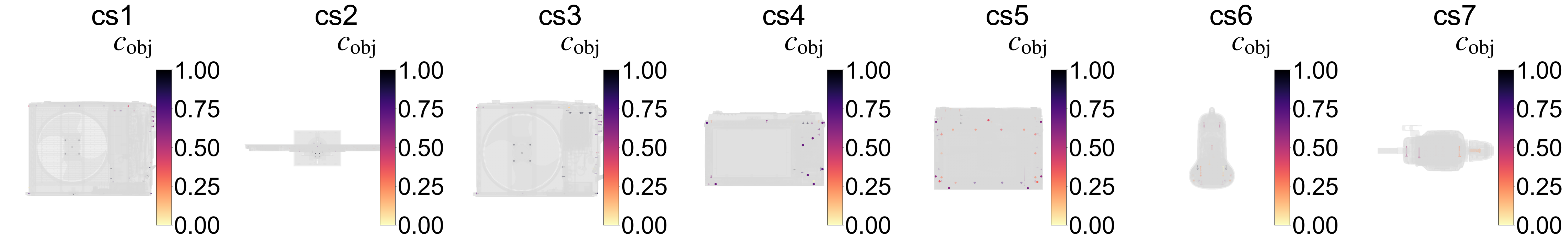}
    \caption{Top-down directive view of $c_\mathrm{obj}$ identifying fasteners whose removal most improves efficiency.}
    \figlab{directive_topview_obj}
\end{figure*}

\subsection{Component Influence Estimation}
To provide a quantitative basis for redesign, the framework calculates influence scores representing the degree to which each part acts as a structural hindrance. Given a feasible baseline robotic disassembly sequence $S$ of length $n$~\cite{Kiyokawa2026schedule}, the system evaluates each target component index $i \in \{1, \dots, n\}$. It generates $n-1$ modified sequences $S_k$ by moving component $i$ to every other position in $S$.

The system counts the resulting structural constraint violations $v_{\mathrm{const}}(S_k)$ and objective-value degradations $v_{\mathrm{obj}}(S_k)$ for each swap sequence $S_k$. The term $v_{\mathrm{const}}$ aggregates disassembly feasibility and stability constraint violations~\cite{Kiyokawa2025,Kiyokawa2026schedule}. The term $v_{\mathrm{obj}}$ aggregates planner evaluation objective degradations. The constraint-violation score $c_{\mathrm{const}, i}$ and the objective-degradation score $c_{\mathrm{obj}, i}$ for component $i$ are the averages over the $n-1$ swaps (Phase 1 of \algoref{full_pipeline}):
\begin{equation} \forlab{c_const}
    c_{\mathrm{const}, i} = \frac{1}{n-1} \sum_{k=1}^{n-1} v_{\mathrm{const}}(S_k)
\end{equation}
\begin{equation} \forlab{c_obj}
    c_{\mathrm{obj}, i} = \frac{1}{n-1} \sum_{k=1}^{n-1} v_{\mathrm{obj}}(S_k)
\end{equation}
The combined influence score $s(i)$ for any component $i$ is defined as the average of these two metrics:
\begin{equation} \forlab{s_score}
    s(i) = \frac{1}{2}\bigl(c_{\mathrm{const}, i} + c_{\mathrm{obj}, i}\bigr)
\end{equation}
The baseline sequence $S$ is a permutation of all $n$ disassembly-target components.

\subsection{Fastener Reduction Simulation}
The framework identifies fastener-reduction candidates by analyzing the CCC graph. Components are identified as fasteners if their node labels $\boldsymbol{I}$ include \texttt{\_screw}, \texttt{\_bolt}, or \texttt{\_nut}. The system scans the graph edge set $E_{CC}$ to group fasteners $\mathcal{F}_p$ attached to the same host part $p$ (Phase 2 of \algoref{full_pipeline}). For each group, the system selects the $r$ fasteners with the highest combined influence score $s(f)$, where $f$ represents the component index of the fastener, as defined in \forref{s_score}. This selection is repeated for each removal count $r \in \{1, \dots, R_\mathrm{max}\}$, where $R_\mathrm{max}$ is the maximum limit. Let $\mathcal{R}$ be the subset of fasteners selected for removal.

To compute the reduction in structural constraints, a binary keep-mask $m$ is applied to each part index. The mask sets $m_j = 0$ if part $j \in \mathcal{R}$, and $m_j = 1$ otherwise. The binary constraint indicator $e_{ij}$ is derived from the constraint-transition matrix $\boldsymbol{X}_{ctm}$. The logical OR operator ($\lor$) ensures that the indicator function $\mathbb{I}$ evaluates to 1 if either part $i$ physically blocks part $j$, or part $j$ blocks part $i$ (Phase 3 of \algoref{full_pipeline}):
\begin{equation} \forlab{e_ij}
    e_{ij} = \mathbb{I}\bigl[\boldsymbol{X}_{ctm}[i,j] \neq 0 \;\lor\; \boldsymbol{X}_{ctm}[j,i] \neq 0\bigr]
\end{equation}
The post-removal constraint count $E_{\mathrm{after}}$ and its reduction $\Delta E$ from the original count $E_{\mathrm{before}}$ are calculated as:
\begin{equation} \forlab{e_after}
    E_{\mathrm{after}}(\mathcal{R}) = \sum_{i<j} m_i m_j e_{ij}
\end{equation}
\begin{equation} \forlab{deltae}
    \Delta E(\mathcal{R}) = E_{\mathrm{after}} - E_{\mathrm{before}}
\end{equation}
The term $m_i m_j$ ensures that a constraint is counted only if both parts physically remain in the assembly, whereas the summation condition $i < j$ limits the calculation to unique component pairs to avoid double counting. To maintain physical validity, an isolated-node check is performed using the contact matrix $\boldsymbol{X}_{ctc}$. This matrix represents physical touching between parts. Any candidate leaving a remaining part completely disconnected from all other parts is discarded as a safety check in \algoref{full_pipeline}.

\subsection{Assessment of Redesign Impacts}
Each candidate is evaluated for geometric stability and robotic operational efficiency. Stability is measured via two geometric stability metrics: the polar moment ratio $\rho_J = J_{\mathrm{after}}/J_{\mathrm{before}}$ and the convex hull area ratio $\rho_A = A_{\mathrm{after}}/A_{\mathrm{before}}$. The term $J$ is the mean squared distance of the remaining fasteners from their geometric center. Both quantities are defined using the individual center coordinates $\boldsymbol{p}_f$ of the remaining fasteners; $J$ additionally uses their arithmetic mean $\boldsymbol{\bar{p}}$:
\begin{equation} \forlab{polar_J}
    J(\mathcal{F}) = \frac{1}{|\mathcal{F}|} \sum_{f \in \mathcal{F}} \lVert \boldsymbol{p}_f - \boldsymbol{\bar{p}} \rVert^2
\end{equation}
\begin{equation} \forlab{area_A}
    A(\mathcal{F}) = \mathrm{Area}\bigl( \mathrm{ConvexHull}_{2}(\{\boldsymbol{p}_f\}_{f \in \mathcal{F}}) \bigr)
\end{equation}
The convex hull $\mathrm{ConvexHull}_{2}$ is computed by projecting points onto a 2D principal-component plane, because a 3D convex hull volume becomes zero for flat layout arrangements. The before and after ratios use $J_{\mathrm{before}} = J(\mathcal{F}_p)$ and $J_{\mathrm{after}} = J(\mathcal{F}_p \text{ excluding } \mathcal{R})$. The area $A$ is handled identically.

Operational efficiency gains are quantified by the reduction in tool changes $\Delta T$ and robot travel distance $\Delta D$. Let $T$ be the count of changes in assigned end-effector $\mathrm{tool}_k$ for step $k$, and let $D$ be the sum of distances between successive components' positions $\boldsymbol{p}$ in sequence $S$:
\begin{equation} \forlab{tool_T}
    T = \sum_{k=1}^{n-1} \mathbb{I}\bigl[\mathrm{tool}_{k+1} \neq \mathrm{tool}_k\bigr]
\end{equation}
\begin{equation} \forlab{dist_D}
    D = \sum_{k=1}^{n-1} \lVert \boldsymbol{p}_{S[k+1]} - \boldsymbol{p}_{S[k]} \rVert
\end{equation}
The reported reductions are $\Delta T = T(S_{\mathcal{R}}) - T(S)$ and $\Delta D = D(S_{\mathcal{R}}) - D(S)$. The term $S_{\mathcal{R}}$ denotes the baseline sequence $S$ excluding the removed fasteners $\mathcal{R}$.

\section{Experiments}

\begin{table*}[!tb]
  \centering
  \caption{Top-ranked fastener-reduction candidates ($R_\mathrm{max} = 3$) compared against a random baseline. Random $\Delta E$, $\Delta T$, and $\Delta D$ report $\mathrm{mean}\pm\mathrm{std}$ over 20 trials; $\rho_J$ and $\rho_A$ report the trial mean. Influence entries are deterministic.}
  \tablab{summary}
    \begin{tabular}{|c|l|c|c|c|c|c|}
    \hline
    \textbf{ID (Product)} & \textbf{Strategy} & $\boldsymbol{\Delta E}$ & $\boldsymbol{\Delta T}$ & $\boldsymbol{\Delta D}$ \textbf{[mm]} & $\boldsymbol{\rho_J}$ & $\boldsymbol{\rho_A}$ \\ \hline
    \multirow{2}{*}{cs1 (Condenser)} & Random (20 trials) & $-7\pm1$ & $0.8\pm1.2$ & $-854\pm585$ & 0.68 & 0.37 \\
     & \textbf{Influence (proposed)} & $-9$ & $-2$ & $-1568$ & 1.06 & 0.00 \\ \hline
    \multirow{2}{*}{cs2 (TV)} & Random (20 trials) & $-7\pm1$ & $0.3\pm0.8$ & $-984\pm740$ & 0.66 & 0.60 \\
     & \textbf{Influence (proposed)} & $-9$ & $0$ & $-1675$ & 0.00 & 1.00 \\ \hline
    \multirow{2}{*}{cs3 (AC Unit)} & Random (20 trials) & $-7\pm1$ & $0.1\pm0.6$ & $-862\pm555$ & 0.64 & 0.50 \\
     & \textbf{Influence (proposed)} & $-9$ & $0$ & $-1109$ & 1.07 & 0.45 \\ \hline
    \multirow{2}{*}{cs4 (Microwave)} & Random (20 trials) & $-7\pm1$ & $0.1\pm0.4$ & $-586\pm402$ & 0.67 & 0.64 \\
     & \textbf{Influence (proposed)} & $-8$ & $0$ & $-462$ & 0.74 & 0.53 \\ \hline
    \multirow{2}{*}{cs5 (Compact MW)} & Random (20 trials) & $-7\pm1$ & $0.2\pm0.6$ & $-613\pm455$ & 0.64 & 0.65 \\
     & \textbf{Influence (proposed)} & $-10$ & $0$ & $-1152$ & 0.00 & 1.00 \\ \hline
    \multirow{2}{*}{cs6 (Beauty Device)} & Random (20 trials) & $-132\pm0$ & $0.0\pm0.0$ & $-151\pm40$ & 1.01 & 0.94 \\
     & \textbf{Influence (proposed)} & $-132$ & $0$ & $-165$ & 1.10 & 1.00 \\ \hline
    \multirow{2}{*}{cs7 (Power Tool)} & Random (20 trials) & $-40\pm6$ & $0.5\pm0.9$ & $-291\pm125$ & 0.94 & 0.67 \\
     & \textbf{Influence (proposed)} & $-36$ & $0$ & $-585$ & 0.62 & 0.44 \\ \hline
    \end{tabular}
\end{table*}

\subsection{Setup}
Seven household appliances are utilized as case studies (cs1--cs7): a condenser unit, a television, an air-conditioner outdoor unit, a microwave oven, a compact microwave oven, an optical beauty device, and a power tool. \figref{objects} shows a 3D model representation of every product. For each product, the proposed analytical framework performs the following steps: (i) constructs the CCC graph, (ii) calculates the influence scores $c_\mathrm{const}$ and $c_\mathrm{obj}$ for each component, (iii) runs the fastener-reduction simulation with $R_\mathrm{max} = 3$, and (iv) evaluates the constraint-reduction, efficiency, and stability metrics ($\Delta E, \Delta T, \Delta D, \rho_J, \rho_A$).

The operational efficiency metrics $\Delta T$ and $\Delta D$ are computed analytically against the two robotic cells illustrated in \figref{robot_cells}: a large and a small dual-arm cell. Both utilize dual-arm robots equipped with multi-view vision sensors, linear and rotary stages to rotate target objects, and end-effector tool racks placed near each arm.

\subsection{Influence Heatmaps}
The framework projects the calculated influence scores onto the 3D geometry to visualize the structural and operational hindrances. \figref{heatmaps_const_overview} and \figref{heatmaps_obj_overview} show the assembly-level heatmaps for $c_\mathrm{const}$ and $c_\mathrm{obj}$, respectively, across all seven products on a normalized $[0, 1]$ scale.

The $c_\mathrm{const}$ heatmaps (\figref{heatmaps_const_overview}) highlight parts whose positions in the robotic disassembly sequence most frequently violate structural constraints. Base chassis parts, such as the condenser unit side covers and the television base plates, are consistently highlighted. This visually confirms that these base parts act as major structural constraints that lock the entire assembly together.

Conversely, the $c_\mathrm{obj}$ heatmaps (\figref{heatmaps_obj_overview}) identify components that degrade the planner's evaluation objectives. Components that violate constraints often simultaneously increase $c_\mathrm{obj}$ by disrupting the overall disassembly flow. Notably, $c_\mathrm{obj}$ tends to be high across exterior housing covers. This occurs because these large components necessitate extensive robot travel and frequent tool changes during the entire disassembly process. By providing both heatmaps, the framework allows engineers to diagnose whether a disassembly difficulty originates from a specific structural interference or a broad operational inefficiency.

To turn these diagnostics into actionable instructions, the framework can also filter the heatmaps to color only the fasteners of multi-fastened groups, rendering the remaining parts as a transparent scaffold. \figref{directive_topview} and \figref{directive_topview_obj} show this directive view for $c_\mathrm{const}$ and $c_\mathrm{obj}$, respectively, across all seven products.

\subsection{Fastener Reduction Recommendations}
\tabref{summary} reports the top fastener-reduction candidates sorted by influence scores. The reduction in structural constraints ($\Delta E$) ranges from 8 to 132 edges; cs6 attains the largest $\Delta E$ because each fastener independently contributes to the structural constraints. The tool-change count ($\Delta T$) decreases by two on cs1. The travel distance ($\Delta D$) is shortened by 165\,mm to 1675\,mm.

To quantify the value of using the influence score as a selection signal, the proposed component-influence rule is compared against a random baseline. This baseline draws $r = 3$ fasteners uniformly at random from each multi-fastened host group, averaged over 20 random trials. The proposed rule matches or exceeds the random baseline on $\Delta E$ for six of the seven products; only on cs7 does random selection attain a slightly larger mean $\Delta E$, as the host group admits several near-equivalent subsets of three fasteners. Tool-change savings ($\Delta T$) also favor the proposed rule: random selection raises $\Delta T$ by an average of $+0.3$, whereas the proposed rule strictly avoids increasing tool changes ($\Delta T \leq 0$). It also yields larger travel-distance savings ($\Delta D$) on six of the seven products. 

However, because high-influence fasteners tend to be concentrated spatially, the proposed rule returns more significant decreases in $\rho_J$ or $\rho_A$ than random selection.

\subsection{Sensitivity to the Removal Limit $R_\mathrm{max}$}
\figref{sensitivity} plots the maximum possible $\Delta E, \Delta T$, and $\Delta D$ as $R_\mathrm{max}$ increases. $\Delta E$ grows monotonically on every product. The marginal gain between $R_\mathrm{max} = 3$ and $4$ is small, indicating that three fasteners is a practical cap. Furthermore, $\Delta T$ saves only when an entire tool-operation run is removed (cs1), and $\Delta D$ likewise saturates at $R_\mathrm{max} = 3$. $R_\mathrm{max} = 3$ is therefore adopted as the default limit.

\begin{figure*}[!tb]
    \centering
    \includegraphics[width=\linewidth]{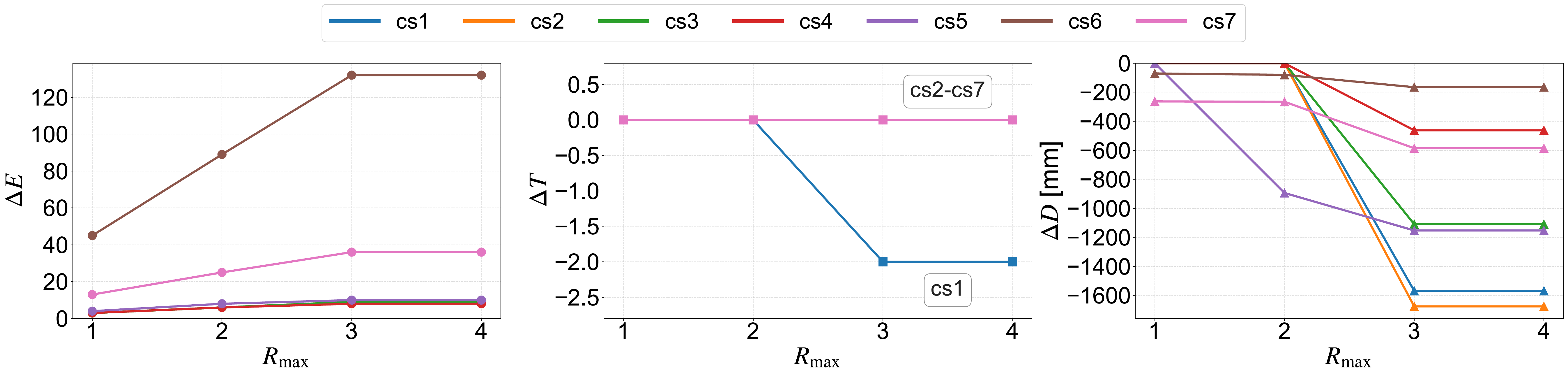}
    \caption{Sensitivity of maximum possible $\Delta E, \Delta T$, and $\Delta D$ to the limit $R_\mathrm{max}$.}
    \figlab{sensitivity}
\end{figure*}

\section{Discussion}
\subsection{Balancing Stability and Efficiency}
The framework evaluates constraint-reduction, efficiency, and stability metrics ($\Delta E, \Delta T, \Delta D, \rho_J, \rho_A$) to prevent structurally unsound modifications. Evaluating only $\Delta E, \Delta T$, and $\Delta D$ could mislead designers into adopting structurally infeasible configurations. The geometric stability metrics ($\rho_J, \rho_A$) reveal the structural risk of each recommendation. Several products (cs1, cs2, cs5) exhibit one ratio collapsing to zero, indicating that the highest-influence fasteners are concentrated on a narrow linear region. Reporting both geometric stability metrics alongside $\Delta E$ enables the designer to reject recommendations whose surviving fastener arrangement is no longer physically sound. Simultaneously assessing geometric stability and robotic operational efficiency satisfies the structural requirements of robotic-friendly design~\cite{Khendry2025}.

\subsection{Generative Redesign Beyond Fastener Reduction}
A consistent pattern across the products is that base chassis parts often contain densely packed fasteners. These fasteners act as major structural constraints that complicate robotic sequences. The top-down directive views (\figref{directive_topview} and \figref{directive_topview_obj}) filter the heatmaps to highlight only fasteners in multi-fastened groups. This visualization surfaces critical fasteners on the top faces or outer perimeters. The $c_\mathrm{obj}$ view (\figref{directive_topview_obj}) instead highlights the fasteners whose removal most improves robotic operational efficiency. 

Currently, designers can reduce and replace these connections by manually substituting multiple small fasteners with fewer robust joints or snap-fits. By leveraging recent advancements in multimodal LLMs and deep generative networks for computer-aided design~\cite{Regenwetter2022,Ma2023,Zhang2026}, this redesign process can be highly automated. Recent studies have demonstrated that generative AI can synthesize complex 3D CAD geometries~\cite{Wu2021,Li2024}, optimize regularized mechanical structures~\cite{Kim2026}, and refine Text-to-CAD models through visual feedback~\cite{Wang2025}.

Future frameworks could feed the quantitative reduction metrics ($\Delta E, \Delta T, \Delta D$) and geometric stability margins ($\rho_J, \rho_A$) extracted by our pipeline directly into these LLM-based CAD systems as spatial bounding constraints and semantic prompts. This integration would enable generative models to automatically synthesize optimized, structurally sound joint structures or snap-fit alternatives exactly at the targeted structural constraints, fully bridging the gap between analytical constraint identification and automated geometric redesign.

\subsection{Physical Fidelity and Estimation Refinement}
The constraint-reduction, efficiency, and stability metrics ($\Delta E, \Delta T, \Delta D, \rho_J, \rho_A$) serve as practical indicators but do not account for material strength or robot dynamics. Future work will integrate finite-element analysis and motion planning while taking advantage of recent advancements in precise contact simulation technologies to provide rigorous validation of cycle-time reductions. Despite the absence of dynamic robot simulations, the current CAD-based analysis effectively identifies and ranks the most critical fasteners for redesign. This demonstrates that structural hindrances can be quantified through geometric relationships alone, providing practical guidance without the need for complex physical modeling of robot motions.

\section{Conclusion}
This paper presented an analytical framework that translates component-level influence scores into ranked fastener-reduction candidates for robotic-disassembly-aware redesign. Candidates are ranked based on their combined influence scores early in the design phase. The proposed pipeline reports changes in $\Delta E$, $\Delta T$, and $\Delta D$, while evaluating geometric stability metrics ($\rho_J, \rho_A$) to prevent unsafe modifications. 

Evaluations conducted on seven household appliances demonstrated that the generated recommendations eliminate between 8 and 132 structural constraints on the graph. Furthermore, they decrease the tool-change count by up to two and shorten the travel distance by 165 to 1675 millimeters. The experimental results further indicated that a maximum removal limit of three fasteners is sufficient. Additionally, this framework allows designers to identify systemic design patterns, enabling the consolidation of redundant fasteners into robust joints early in the design phase.

Future research will focus on replacing these simplified geometric approximations with high-fidelity finite-element analysis, validating efficiency metrics through integrated motion planning, and expanding the scope of design recommendations to include part translation, rotation, and snap-fit alternatives.

\section*{Acknowledgement}
This work was supported by the New Energy and Industrial Technology Development Organization (NEDO) project JPNP23002.

\bibliographystyle{IEEEtran}
\bibliography{refdata}

\end{document}